\documentclass[letterpaper, 10 pt, conference]{URL-ieeeconf}
\IEEEoverridecommandlockouts      
\usepackage{graphics}           
\usepackage{times}              
\usepackage{amsmath}            
\usepackage{amssymb}            
\usepackage{graphicx}
\usepackage{tabularx}
\usepackage{booktabs}
\usepackage{algorithm}
\usepackage[noend]{algpseudocode}
\usepackage{booktabs}
\usepackage{color}
\usepackage{multirow}
\usepackage{subcaption}
\usepackage{rotating}
\usepackage{cite}
\usepackage{booktabs}
\usepackage{multirow}
\usepackage{graphicx}
\usepackage[table,xcdraw]{xcolor}
\usepackage{tikz}
\usetikzlibrary{tikzmark}
\definecolor{instructioncolor}{rgb}{.5,.5,.5}
\let\labelindent\relax
\usepackage{enumitem}

\def\figref#1{Fig.~\ref{#1}}

\def\eqref#1{(\ref{#1})}

\newcommand{\rom}[1]{\uppercase\expandafter{\romannumeral #1\relax}}

\makeatletter
\usepackage{xspace}
\DeclareRobustCommand\onedot{\futurelet\@let@token\@onedot}
\def\@onedot{\ifx\@let@token.\else.\null\fi\xspace}

\makeatother

\usepackage{array}
\newcolumntype{L}[1]{>{\raggedright\let\newline\\\arraybackslash\hspace{0pt}}m{#1}}
\newcolumntype{C}[1]{>{\centering\let\newline\\\arraybackslash\hspace{0pt}}m{#1}}
\newcolumntype{R}[1]{>{\raggedleft\let\newline\\\arraybackslash\hspace{0pt}}m{#1}}

\usepackage{tabularx}
\usepackage{booktabs}
\usepackage{makecell}
\usepackage[colorlinks=true, linkcolor=black, citecolor=green, urlcolor=blue]{hyperref}

\title{\LARGE \bf CLUE: Adaptively Prioritized Contextual Cues by Leveraging a Unified Semantic Map for Effective Zero-Shot Object-Goal Navigation
}
\author{Taeyun Kim$^{1}$, Alvin Jinsung Choi$^{1}$, Dasol Hong$^{1}$ and Hyun Myung$^{1*}$, \textit{Senior Member, IEEE}
  \thanks{$^*$Corresponding author: Hyun Myung}
  \thanks{All authors are with the School of Electrical Engineering, KAIST (Korea Advanced Institute of Science and Technology), Daejeon, 34141, Republic of Korea. {\ttfamily\scriptsize \{ktw1404, alvinjinsung, ds.hong, hmyung\}@kaist.ac.kr} \hfill \break       
     }
}

\begin{document}
\maketitle
\thispagestyle{empty}
\pagestyle{empty}


\begin{abstract}
Zero-shot object-goal navigation (ZSON) is a challenging problem in robotics that requires a comprehensive understanding of both language and visual observations. Contextual cues from rooms and objects are critical, but their relative importance depends on the target: some objects are strongly tied to specific room types, while others are better predicted by nearby co-located objects. Existing methods overlook this distinction, leading to inefficient and inaccurate exploration. We present CLUE, a novel navigation framework that adaptively balances the use of contextual rooms and objects by leveraging commonsense knowledge extracted from an offline large language model (LLM). By estimating a target’s association with room types using LLM, the agent prioritizes room cues for predictable objects and object cues for those with weak room associations. Our framework constructs a unified semantic value map that integrates both types of contextual information, adaptively weighted by the target’s ambiguity to guide exploration. Combined with multi-viewpoint verification and an exploration strategy informed by contextual cues, CLUE achieves robust and efficient navigation. Extensive experiments in simulation and real-world deployments show that our method consistently outperforms state-of-the-art baselines in both success rate (SR) and success weighted by path length (SPL), demonstrating its effectiveness and practicality for real-world navigation tasks.

\end{abstract}

\section{Introduction}
\label{sec:intro}

\newcommand{\TBD}{{CLUE}}

Navigating to find a target object in an unseen environment, known as object-goal navigation (ObjectNav)~\cite{batra2020objectnav, sun2024survey}, is a fundamental challenge in robotics and embodied AI. Success in this task requires more than accurate object detection. The agent must reason about where objects are likely to appear by leveraging the environmental context. For example, the agent needs to recognize that toilets are usually located in bathrooms, while chairs are often found near tables. This requires linking object categories to their typical spatial and functional contexts, enabling the agent to infer potential object locations and bridge the gap between abstract commands and physical exploration.

Recent advances in foundation models such as large language models (LLMs)~\cite{achiam2023gpt, touvron2023llama} and vision-language models (VLMs)~\cite{radford2021learning, li2023blip} have enabled rapid progress in zero-shot ObjectNav (ZSON)~\cite{majumdar2022zson, gadre2023cows, vlfm, busch2024one, zhang2025apexnav, zhou2023esc, sg-nav, wu2024voronav, kuang2024openfmnav, zhong2024topv, yu2023l3mvn, zhang2024trihelper}. ZSON offers a more general and efficient setting for real-world deployment, as it removes the need for task-specific training and allows an agent to adapt to new targets and environments on the fly. Some works~\cite{gadre2023cows, majumdar2022zson} employed CLIP~\cite{radford2021learning} to align visual inputs with target categories, enabling zero-shot recognition of objects in novel scenes. Subsequent approaches~\cite{vlfm, busch2024one, zhang2025apexnav} extended this idea by integrating similarity between the current visual observation and the target category into a physical semantic value map, supporting better spatial reasoning. Other works~\cite{sg-nav, wu2024voronav, zhong2024topv, yu2023l3mvn, kuang2024openfmnav, zhang2024trihelper} leveraged LLMs for higher-level reasoning, using them to interpret accumulated observations and plan subsequent actions. Collectively, these efforts have established a strong foundation for ZSON by integrating powerful vision–language understanding with spatial reasoning.



\begin{figure}[!t]
    \centering
    \captionsetup{font=footnotesize}
    \includegraphics[width=1.0\linewidth]{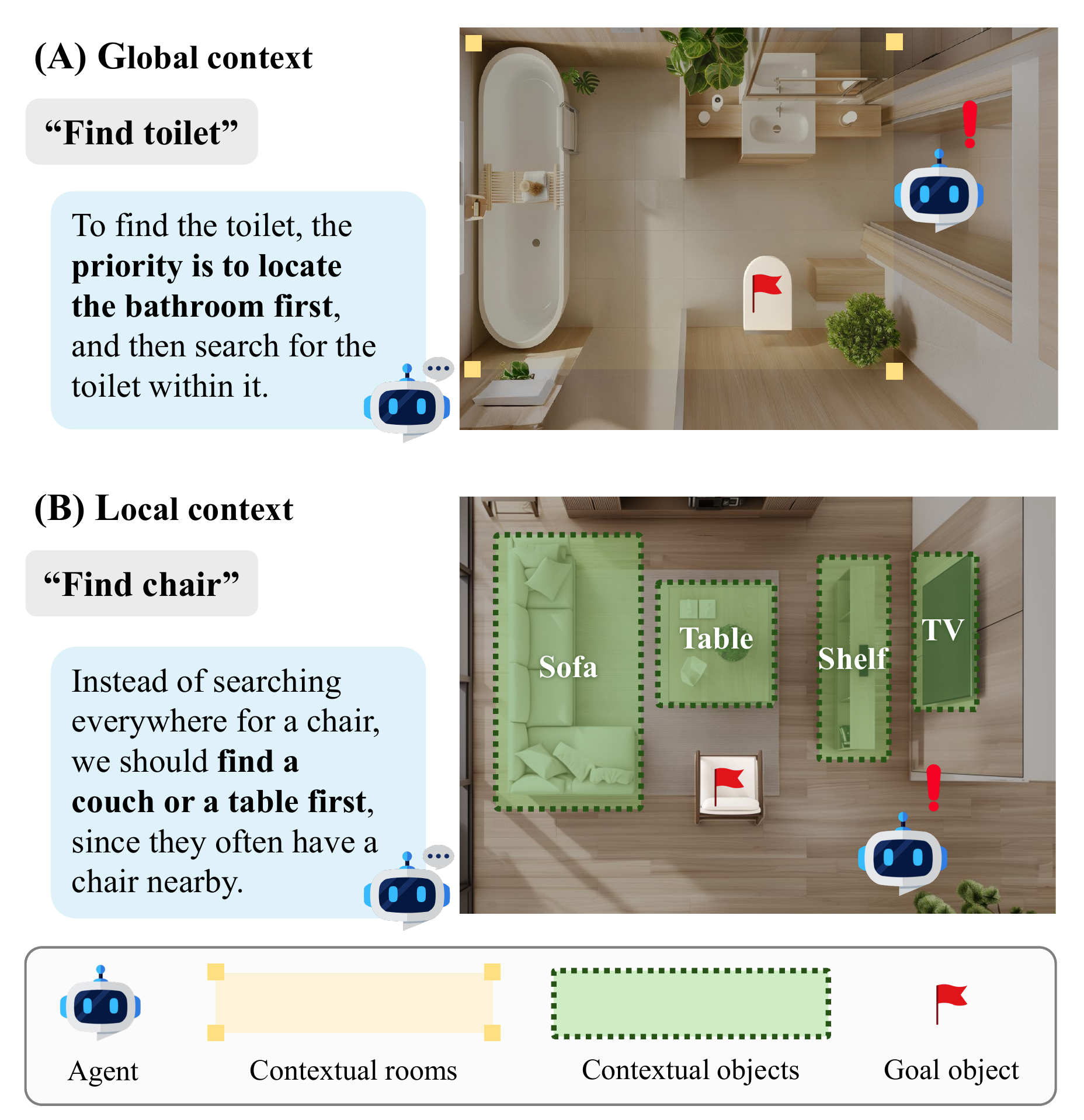}
    \caption{
    Illustration of our adaptive strategy for target object search. For objects strongly correlated with particular room types, the agent emphasizes global context using contextual rooms where the target is likely to appear. For objects with low association to specific room types, the agent emphasizes local context using contextual objects that are likely to be co-located with the target.
    }
    \label{fig:1}
    \vspace{-10pt}
\end{figure}


Despite recent progress, current ZSON methods do not adaptively adjust the relative importance of contextual cues, limiting their effectiveness and accuracy. In many cases, contextual rooms where an object is typically found, and contextual objects it is often co-located with, provide strong semantic cues for the ObjectNav. Although both cues are valuable, their relative importance depends on the target. For example, when searching for a toilet that is strongly associated with a bathroom, prioritizing contextual rooms enables more efficient navigation. Conversely, when searching for a chair that is not tied to a specific room type, emphasizing contextual objects such as a table leads to reliable navigation. In summary, contextual rooms provide a stronger signal for objects strongly tied to a specific room type, whereas contextual objects should be emphasized for objects whose locations can be inferred from co-located items.

We propose \textbf{\TBD}~(Adaptively Prioritized \textbf{C}ontextual Cues by \textbf{L}everaging a \textbf{U}nified Semantic Map for \textbf{E}ffective Zero-Shot Object-Goal Navigation), a novel navigation framework for ZSON that constructs a unified semantic value map with adaptively prioritized contextual cues. We first quantify the value of each cue. The target object score serves as a base for our semantic value map, which utilizes a vision-language model (VLM) to compute the similarity between the current observation and the target object category. The contextual room score is obtained using a similar technique, with the text prompt changed to predefined room types. The contextual object score is computed from the semantic correlation between the target and contextual objects and is modeled as a Gaussian distribution centered on their locations. These scores are then fused into a unified semantic value map, weighted by the normalized entropy of the target object's room association probability distribution, derived from offline LLM queries. This adaptive weighting allows the semantic value map to reflect the characteristics of the target object by balancing global room-level cues and local object-level cues, enabling more effective navigation. For targets with low entropy on room associations (Fig.~\ref{fig:1}(a)), the semantic value map prioritizes contextual room scores, highlighting global environmental cues. In contrast, for targets with high entropy (Fig.~\ref{fig:1}(b)), greater weight is assigned to contextual object scores, emphasizing local spatial relationships.

The agent navigates using this weighted semantic value map by sequentially moving toward the most promising frontier. To ensure robustness, \TBD~performs multi-viewpoint observations of each target candidate for robust verification. A further advantage of our framework is its ability to perform higher-level reasoning without relying on costly online LLM queries. Instead, we use offline LLM queries to extract commonsense knowledge about target–room associations, contextual rooms, and co-located objects before execution, which significantly reduces computational overhead and avoids delays during real-world navigation.

We validate our approach through extensive experiments on the HM3D~\cite{ramakrishnan2021habitat} dataset. \TBD~achieves state-of-the-art performance, outperforming competitive baselines in both success rate (SR) and success weighted by path length (SPL). We further deploy \TBD~on a Clearpath Jackal platform in real-world settings, demonstrating its practicality for on-board navigation.
 
In summary, our work makes three key contributions:

\begin{itemize}

\item We introduce \textbf{\TBD}, a framework that constructs a unified semantic value map by balancing contextual room and contextual object cues according to the target’s characteristics, enabling more reliable exploration.

\item We ensure real-time capability by leveraging offline LLM queries for commonsense knowledge, eliminating the latency and computational overhead of online LLM reasoning.

\item We validate the effectiveness of our method through extensive experiments in both simulation benchmark and real-world deployment, achieving state-of-the-art performance compared with state-of-the-art baselines.

\end{itemize}
\section{Related Works}
\label{sec:related}


\subsection{Zero-Shot Object-Goal Navigation}

With access to internet-scale training data, LLMs~\cite{achiam2023gpt, touvron2023llama} and VLMs~\cite{radford2021learning, li2023blip} have demonstrated strong generalization and reasoning capabilities. Recent works~\cite{gadre2023cows, vlfm, busch2024one, zhang2025apexnav} have leveraged these models to tackle object-goal navigation in a zero-shot setting. COWs~\cite{gadre2023cows} employs CLIP~\cite{clip} to align the current visual observation with the target object description, but relies on single-frame signals that lack spatial-temporal context. VLFM~\cite{vlfm} constructs a semantic value map by computing CLIP-based similarity between the target object and current observations for better reasoning, but uses the same strategy for all targets without distinguishing whether they are better localized through room cues or co-located objects. Onemap~\cite{busch2024one} extends VLFM to sequential multi-object navigation by maintaining a memory of prior search locations, and ApexNav~\cite{zhang2025apexnav} builds on VLFM by adapting exploration based on the magnitude of semantic cues. However, both methods also employ an object-agnostic strategy, thereby preventing the full exploitation of contextual information.

While these methods demonstrate the potential for ZSON, they do not fully leverage contextual cues. Most rely only on local information or do not consider the relative importance of cues. In contrast, our approach constructs a unified semantic value map that adaptively balances global room-level and local object-level cues, weighting them according to the relative importance based on the target.


\subsection{LLM-based Reasoning for ObjectNav}

Recent works~\cite{zhou2023esc, sg-nav, wu2024voronav, kuang2024openfmnav} leverage LLMs for higher-level reasoning by incorporating commonsense knowledge. These methods provide the LLM with structured representations of the scene, transforming raw observations into formats that capture object relationships and spatial context to improve decision-making. ESC~\cite{zhou2023esc} queries an LLM to infer which frontiers may lead to the target based on detected objects. SG-Nav~\cite{sg-nav} constructs a hierarchical 3D scene graph and queries the LLM to select navigation actions. VoroNav~\cite{wu2024voronav} encodes a topological map together with semantic information into textual descriptions of the environment and exploratory paths, which are then provided to the LLM to determine actions. OpenFMNav~\cite{kuang2024openfmnav} builds a semantic score map from human instructions and employs LLM-based commonsense reasoning to guide exploration. However, these methods rely on online LLM queries, which often introduce computational overhead and redundancy.

Although LLMs provide strong reasoning capabilities through commonsense knowledge, continuously querying them during execution hinders real-world applicability due to high latency. In contrast to other methods, we leverage LLMs offline to extract the most relevant knowledge about the target object prior to execution. This strategy avoids runtime delays while preserving high-level reasoning ability, enabling efficient navigation without redundant online queries.




\begin{figure*}[!t]
    \centering
    \captionsetup{font=footnotesize}
    \includegraphics[width=1.0\linewidth]{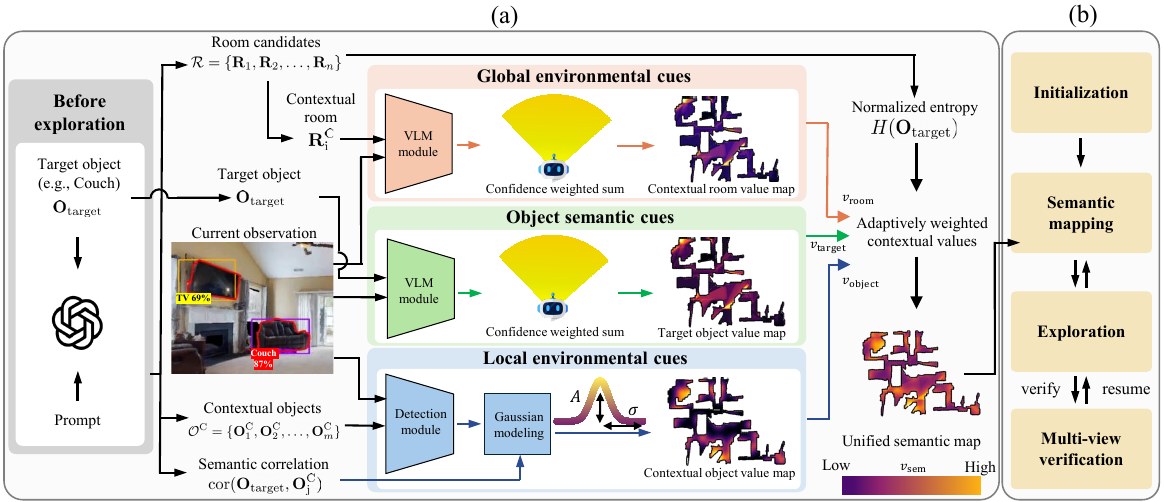}
    \caption{
    Overview of \TBD. (a) Construction of a unified semantic value map by adaptively balancing contextual cues according to the target’s characteristics. Prior to execution, commonsense knowledge is extracted from an LLM, including contextual rooms with their associated probabilities and contextual objects with their semantic relevance to the target. This information is used to build contextual room and contextual object value maps, which are then fused adaptively using entropy-based weighting to capture global and local environmental cues. (b) Utilization of the unified semantic value map for exploration through a coarse-to-fine strategy, where semantically relevant regions are prioritized and multi-view verification ensures efficient and accurate navigation.
    }
    \label{fig:overview}
\vspace{-5pt}
\end{figure*}

\section{\TBD: Unified Semantic Value Map for Scene Understanding}
\label{sec:main}

\newcommand{\contextroom}[1]{\mathbf{R}_{#1}^\text{C} }
\newcommand{\contextobject}[1]{\mathbf{O}_{#1}^\text{C} }
\newcommand{\target}{\mathbf{O}_{\text{target}} }
\newcommand{\room}[1]{\mathbf{R}_{#1}}
\newcommand{\setroom}{\mathcal{R} }
\newcommand{\setcontextroom}{\mathcal{R}^\text{C} }
\newcommand{\setcontext}{\mathcal{O}^\text{C} }
\def\figref#1{Fig.~\ref{#1}}

\TBD~represents the scene with a unified semantic value map that integrates contextual room and object cues, weighted by their relative importance to the target for reliable navigation. Section~\ref{sec:problem_definition} defines the problem, and Section~\ref{sec:map} introduces the map representation that provides the spatial foundation for expressing contextual cues and guiding navigation. Section~\ref{sec:target_object} details how target object cues are quantified as scores, which serve as the basis for the semantic map. Section~\ref{sec:contextual_room} and Section~\ref{sec:contextual_object} explain how contextual rooms and objects are identified and scored. Finally, Section~\ref{sec:semantic_map} describes how these signals are fused into a unified semantic value map with adaptive weighting. The overall framework is illustrated in Fig.~\ref{fig:overview}.


\subsection{Problem Definition}
\label{sec:problem_definition}

In ObjectNav, an agent is placed in an unknown environment with no prior map and tasked to locate a target object specified by a natural language query (e.g., `bed' or `chair'). The agent’s perception is limited to an egocentric camera view. A trial is considered successful if the agent stops within the distance of $d_s$ from the target object in at most $T$ steps.


\subsection{Map Representation}
\label{sec:map}

We construct a top-down 2D map to represent the environment. Similar to VLFM~\cite{vlfm}, we employ a 2D value map composed of a geometric map $\mathcal{M}_{\text{geo}}$ and a semantic value map $\mathcal{M}_{\text{sem}}$ to represent the environment.

The geometric map $\mathcal{M}_{\text{geo}}$ is generated by projecting the 3D point cloud from an RGB-D camera or LiDAR into a 2D plane, identifying traversable areas, obstacles, and unexplored regions. Local maps are continuously integrated into a global frame using the robot’s pose, classifying regions as explored, unexplored, or occupied. Frontiers are defined as center points along boundaries between explored and unexplored regions to guide exploration.

The semantic value map $\mathcal{M}_{\text{sem}}$ provides a human-like interpretation of visual information, capturing the environment’s semantic structure. Each cell encodes the strength of contextual cues, indicating the likelihood of the target’s presence and guiding navigation. Values are computed by incorporating not only the target object cues but also contextual room and object cues.


\subsection{Target Object Cues}
\label{sec:target_object}

\TBD~aims to locate the target object in an unknown environment by quantifying target object cues as the basis of the semantic value map. Following VLFM\cite{vlfm}, we leverage a VLM to compute target object scores. In particular, BLIP2~\cite{li2023blip} is used to calculate $v_{\text{target}}$ as the cosine similarity between the current view and the target category. This score represents the relevance of the view to the target and is stored in the corresponding pixel of the value map, which is continuously updated as new images are acquired.

Semantic confidence $c$ controls how the target object value of a previously observed pixel is updated when revisited. It is defined as follows:

\begin{equation}
\label{eq_confidence_score}
    c = \cos^2{\left(\frac{\theta}{\theta_{\text{FoV}}/2} \times \frac{\pi}{2}\right)},
\end{equation}
where $\theta$ is the angle between the pixel and the optical axis, and $\theta_{\text{FoV}}$ is the camera's horizontal field of view (FoV). The confidence approaches $1$ for pixels near the image center and decreases toward $0$ at the edges. This reflects that the cosine similarity is generally higher in central regions. Note that $c$ affects only the update of previously observed pixels, not the assignment of values to newly observed ones.

The target object score is updated based on the confidence score, together with current and previous values. This update is performed through a weighted average of the current and previous values and confidence scores as follows:

\begin{align}
    \label{eq_vlm_update}
    v_{\text{target}}^{\text{new}} &= \frac{c^{\text{cur}} v_{\text{target}}^{\text{cur}} + c^{\text{prev}} v_{\text{target}}^{\text{prev}}}{c^{\text{cur}} + c^{\text{prev}}}, \\
    \label{eq_confidence_update}
    c^{\text{new}} &= \frac{\left(c^{\text{cur}}\right)^2 + \left(c^{\text{prev}}\right)^2}{c^{\text{cur}} + c^{\text{prev}}},
\end{align}
where $v_{\text{target}}^{\text{new}}$ and $c^{\text{new}}$ denote the updated target object value and confidence score; $v_{\text{target}}^{\text{cur}}$ and $c^{\text{cur}}$ come from the current observation; and $v_{\text{target}}^{\text{prev}}$ and $c^{\text{prev}}$ are those stored from the previous step.


\subsection{Contextual Rooms for Global Spatial Understanding}
\label{sec:contextual_room}

Contextual rooms strongly associated with the target object provide valuable cues for global spatial reasoning. Given a predefined set of common indoor room categories $\setroom = \{\room{1}, \room{2}, \dots, \room{n}\}$, we query an LLM to identify the room in which the target object $\target$ is most likely to be found as contextual room $\contextroom{i}$. Our goal is to identify a contextual room and leverage it to guide navigation toward the target object.

To this end, we compute the contextual room score $v_{\text{room}}$ in the same manner as the target object score $v_{\text{target}}$, but replace the VLM text prompt with the identified contextual room category. Thus, $v_{\text{room}}$ reflects the relevance of the current view to the most relevant room, emphasizing global spatial cues. The confidence measure for each pixel is defined in the same way as for $v_{\text{target}}$, and the value map is updated using the same weighted update rule.


\subsection{Contextual Objects for Local Spatial Understanding}
\label{sec:contextual_object}

Contextual objects that are often co-located with the target object serve as meaningful cues for local spatial reasoning. To identify them, we first query the LLM for a set of contextual objects $\setcontext = \{\contextobject{1}, \dots, \contextobject{m} \}$ that are semantically related to the target object $\target$. We then compute the semantic correlation $\text{cor}(\target, \contextobject{j})$ between the target object and each contextual object, normalizing the scale so that the target’s self-correlation is equal to $1.0$. The correlations with contextual objects are precomputed by the LLM before the task begins. During exploration, these values are used by the agent to assign contextual object scores when such objects are detected by the detection module.

We employ a detection module to recognize contextual objects and compute the contextual object score $v_{\text{object}}$, modeled as a Gaussian distribution. This score reflects the likelihood of the target object’s presence, being higher near the detected contextual objects and decaying smoothly with distance. For each detection, the module records the image pose, the center of the object point cloud cluster $(p_x, p_y)$, the object class, and the detector's confidence score as an object node. The object cluster center is then projected using the point cloud to obtain $(\bar{x}, \bar{y})$ coordinates on the 2D semantic value map $\mathcal{M}_{\text{sem}}$. At this location, the precomputed correlation score, $\text{cor}(\target, \contextobject{j})$, is utilized to score the pixel and the surrounding area. The contextual object score $v_{\text{object}}$ at 2D map location $(x, y)$ is defined as follows:
\vspace{-4pt}

\begin{equation} 
\label{eq:v_context}
    \begin{aligned}
        &v_{\text{object}}(x, y)\\
        &= A(\target, \contextobject{j}) \cdot \, \exp{\left(-\frac{(x-\bar x)^2 + (y-\bar y)^2}{2\sigma(\target, \contextobject{j})^2}\right)},
    \end{aligned}
\end{equation}
where $A(\target, \contextobject{j})$ is the amplitude that reflects the importance of the contextual object, and $\sigma(\target, \contextobject{j})$ controls the spatial spread of the contextual influence around the detected object’s position $(\bar{x}, \bar{y})$. Formally, it is defined as follows:
\vspace{-4pt}
\begin{align}
    A(\target, \contextobject{j}) &= A_{0} \cdot \text{cor}(\target, \contextobject{j}), \\
    \sigma(\target, \contextobject{j}) &=  \sigma_{0} + \text{cor}(\target, \contextobject{j}),
\end{align}
where $A_{0}$ and $\sigma_{0}$, denote the base amplitude and the base standard deviation, respectively. In this design, stronger semantic correlation not only increases the weight of contextual objects through a larger amplitude but also broadens the spatial influence via a wider spread, enabling the agent to leverage co-located objects more effectively.


\subsection{Fusing Contextual Cues for Unified Semantic Value Map}
\label{sec:semantic_map}


We fuse target object scores along with contextual room and object scores into a unified semantic value map, adaptively weighting cues according to the target’s characteristics. The relative importance is determined by the association of the target object with specific room types. To estimate this association, we query an LLM to obtain the probability $P(\room{k}|\target)$ that the target object is likely to exist in the room type $\room{k}$. From this probability distribution, we then compute the information entropy as a measure of the uncertainty in the object’s candidate room location as follows:

\begin{equation}\label{eq:semantic_entropy} 
    H(\target) = \frac{ - \sum_{k=1}^{n} P(\room{k}|\target)\text{log}( P(\room{k}|\target))}{\text{log}(n)}.
\end{equation}

\noindent We provide the agent with a comprehensive semantic representation to guide the search for the target object $\target$. The target object score $v_{\text{room}}$, the contextual room score $v_{\text{room}}$, and the contextual object score $v_{\text{object}}$ are integrated into the semantic value map $\mathcal{M}_{\text{sem}}$, with each cell value given by:

\begin{equation} \label{eq:total_v}
    v_{\text{sem}} = v_{\text{target}} + \omega_{\text{room}} \cdot v_{\text{room}} + \omega_{\text{object}} \cdot v_{\text{object}},
\end{equation}
where the weights $\omega_{\text{room}}$ and $\omega_{\text{object}}$ are determined by the entropy $H(\target)$. It is formally defined as follows:
\begin{equation}\label{eq:contextual_weights}
    \begin{aligned}
    \omega_{\text{room}} &= 1 - H(\target) \\
    \omega_{\text{object}} &= H(\target)~.\\
   \end{aligned}
\end{equation}



\begin{figure}[!t]
    \centering
    \captionsetup{font=footnotesize}
    \includegraphics[width=1.0\linewidth]{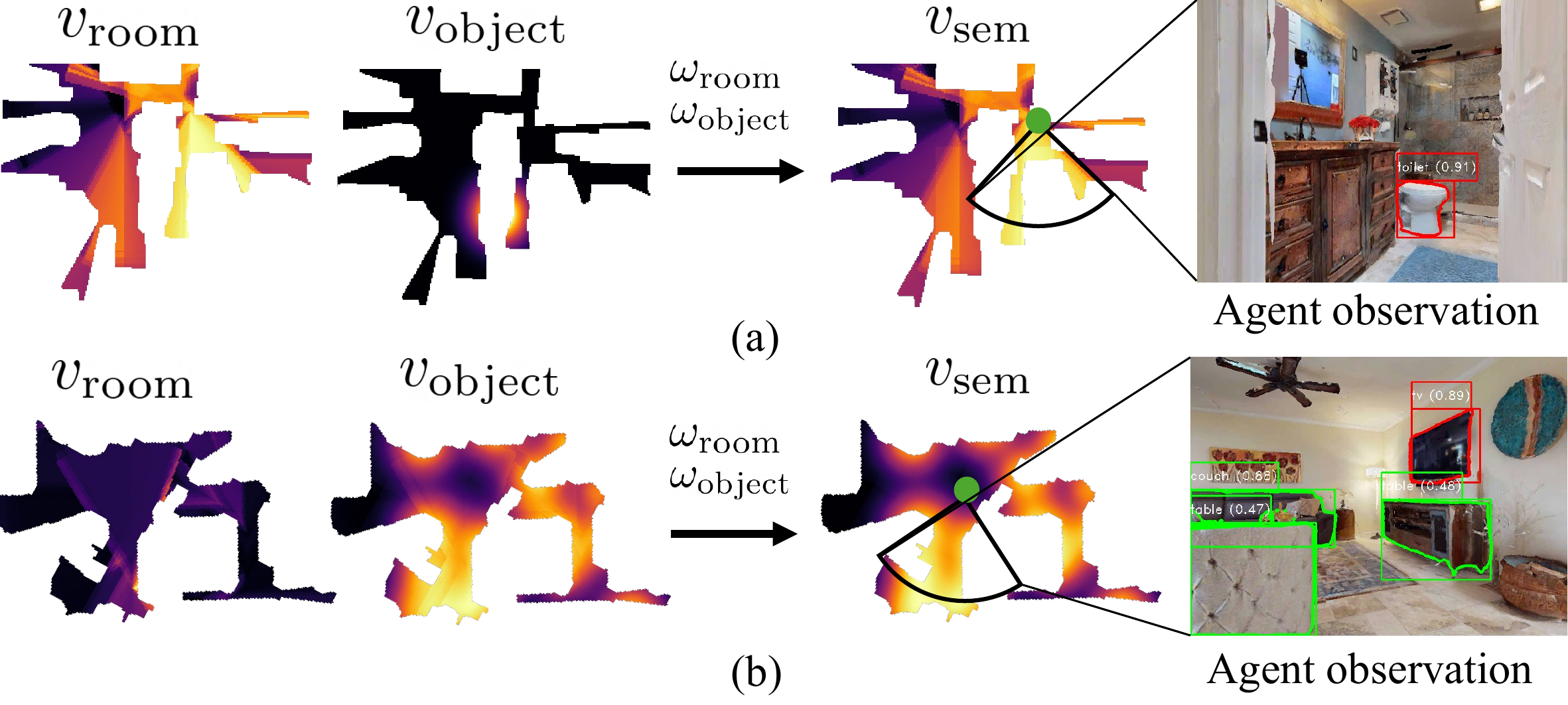}
    \caption{
    Illustration of the agent’s current observation and its corresponding representation on the individual value maps ($v_\text{room}$, $v_\text{object}$) and the final unified semantic value map $v_\text{sem}$. (a) An example of a low-entropy object (toilet), where contextual rooms provide distinctive guidance while contextual objects do not. (b) An example of a high-entropy object (TV), where the unified map is more strongly influenced by local contextual objects due to the lack of distinctive spatial evidence from contextual rooms.
    }
    \label{fig:weight}
    \vspace{-10pt}
\end{figure}


Through this adaptive modeling, low-entropy target objects that are strongly associated with a specific room category prioritize contextual room cues over contextual object cues, emphasizing global environmental signals. In contrast, for high-entropy target objects that may appear across diverse room types, we emphasize contextual object cues as they provide the most reliable and discriminative information for guiding navigation, highlighting local environmental signals. As shown in Fig.~\ref{fig:weight}, the semantic value map captures the spatial properties most relevant to the target object and effectively guides the agent’s navigation.


\section{\TBD: Exploration Strategy}

 
\subsection{Initialization}

At the start of navigation, an agent with a limited FoV has insufficient semantic cues to reliably guide its movement. To mitigate this, the agent first performs an in-place rotation to capture the maximum possible observations from its initial position. Once it has accumulated sufficient semantic and geometric understanding of the surroundings, the agent selects the most promising frontier as its destination and begins moving toward it.


\subsection{Exploration Using Fused Semantic Value Map}

We adopt frontier-based exploration as our navigation policy. The agent selects the frontier it considers the most likely to lead to the target object. To improve efficiency, we incorporate the strategy proposed in ApexNav~\cite{zhang2025apexnav}. 
We first determine if sufficient semantic cues have been accumulated, indicated by checking whether any frontier score on the unified semantic value map $\mathcal{M}_{\text{sem}}$ exceeds a predefined threshold. Before sufficient semantic information is provided, the agent performs geometry-based exploration by solving a traveling salesman problem (TSP) to plan its path. After the condition is met, the agent switches to selecting the frontier with the highest semantic score as its navigation goal. This results in a conditional exploration strategy in which the agent performs rapid geometry-driven exploration when semantic information is sparse, then transitions to semantic-driven exploration once reliable cues become available.





\subsection{Multi-View Verification}

Once a potential target is identified, the agent enters a verification phase. Standard single-image detection can be prone to occlusion and false positives, so our method navigates to multiple viewpoints around the candidate. The detection results from these viewpoints are aggregated, taking into account both the number of consistent detections and their class agreement. Based on this evidence, the system determines whether the candidate is a true target or a false positive. If deemed a false positive, the agent discards the candidate and resumes the search.

\section{Experimental Setting for Evaluation}
\label{sec:exp_set}


\subsection{Dataset}

We evaluate \TBD~by comparing our proposed method with state-of-the-art methods on the ObjectNav task in the Habitat simulator~\cite{savva2019habitat} using the HM3D dataset~\cite{ramakrishnan2021habitat}. The HM3D dataset contains $2{,}000$ episodes across $20$ scenes with $6$ types of target objects. It features diverse, building-scale environments, making it suitable for demonstrating the effectiveness and robustness of the proposed framework.


\subsection{Baselines}

We categorize the baselines based on how they enable the agent to interpret unknown environments. End-to-end methods jointly learn environment representations and ObjectNav policies, including ZSON~\cite{majumdar2022zson}, SemExp~\cite{semexp}, and PONI~\cite{poni}. Although ZSON is often regarded as zero-shot, it still requires ImageNav training. VLM-based methods~\cite{gadre2023cows, vlfm, zhang2025apexnav} compare the target with the current observation, with VLFM~\cite{vlfm} and ApexNav~\cite{zhang2025apexnav} notable for its value map formulation similar to ours. LLM-based methods such as SG-Nav~\cite{sg-nav} and VoroNav~\cite{wu2024voronav} build structured scene representations and query an LLM for action decisions. We also include other LLM-driven approaches~\cite{zhou2023esc, zhong2024topv, yu2023l3mvn, kuang2024openfmnav, zhang2024trihelper} that model complex environmental relationships. All baselines are evaluated in the same environments using both success rate and efficiency.


\subsection{Metrics}

We use two standard metrics~\cite{anderson2018evaluation} to evaluate ObjectNav performance: success rate (SR) and success weighted by path length (SPL). SR is the proportion of episodes in which the agent successfully reaches the target object. SPL measures navigation efficiency by comparing the agent’s actual path length to the shortest possible path length to the target.


\subsection{Implementation Details}

For each episode, the maximum number of allowed steps is set to $T = 500$, with a success distance threshold of $d_s = 0.2\text{m}$. The agent is equipped with an RGB-D sensor providing images at a resolution of $640 \times 480$, with depth values ranging from $0.5\text{m}$ to $5.0\text{m}$. The camera is mounted $0.88\text{m}$ above the ground. The agent performs discrete actions consisting of moving forward by $0.2\text{m}$ or rotating by $30^\circ$. For the VLM used to compute the contextual room score, we employ the BLIP2~\cite{li2023blip}. The Gemini-Pro 2.5 model~\cite{gemini} is used to obtain the semantic entropy $H(\target)$, identify contextual objects, and compute $\text{cor}(\target, \contextobject{j})$. For object detection, we use YOLOv7~\cite{wang2023yolov7} for COCO~\cite{cocodata} classes and GroundingDINO~\cite{liu2024grounding} for non-COCO classes. For segmentation, we use MobileSAM~\cite{mobsam}. All experiments are conducted on RTX 3090 GPUs.

For real-world experiments, we deploy \TBD~on the Clearpath Jackal platform with a Velodyne VLP-16 LiDAR and a Ricoh Theta Z1. To process the omnidirectional sensor data within our framework, the observation area is limited to 90° quadrants. Due to the sparsity of the point cloud, TRIP~\cite{oh2024trip} and TRG-planner~\cite{lee2025trg} were used to extract ground planes and plan path for safe navigation.

\section{Experiment Results}
\label{sec:exp_result}




\setlength{\tabcolsep}{12pt}
\begin{table}[t!]
    \captionsetup{font=footnotesize}
    \centering
    \caption{
    Performance comparison of diverse methods on the HM3D benchmark. Bold indicates the best performance, and gray highlighting indicates the second-best. (a) End-to-end methods. (b) Zero-shot ObjectNav methods.
    }
    \label{table1:overall}
    \begin{tabularx}
        {\columnwidth}{@{}llccc} 
        \toprule
        \midrule
        \multicolumn{2}{c}{\multirow{2}{*}[-0.5ex]{Method}} & 
        \multirow{2}{*}[-0.5ex]{LLM query} & 
        \multicolumn{2}{c}{HM3D} \\
        \cmidrule(lr){4-5}
        & & & SR↑ & SPL↑ \\
        \midrule
        \parbox[t]{1mm}{\multirow{1}{*}{(a)}}
        & ZSON~\cite{majumdar2022zson}        & -  & 25.5 & 12.6  \\
        \midrule
        \parbox[t]{2mm}{\multirow{10}{*}{(b)}}
        & VLFM~\cite{vlfm}                    & -  & 52.5 & 30.4  \\
        & ApexNav~\cite{zhang2025apexnav}     & Offline  & \cellcolor{gray!30}59.6  & \cellcolor{gray!30}33.0   \\
        & ESC~\cite{zhou2023esc}              & Online  & 39.2 & 22.3 \\
        & VoroNav~\cite{wu2024voronav}        & Online & 45.0 & 26.0  \\
        & TopV-Nav~\cite{zhong2024topv}       & Online & 45.9 & 28.0   \\
        & L3MVN~\cite{yu2023l3mvn}            & Online & 50.4 & 23.1  \\
        & OpenFMNav~\cite{kuang2024openfmnav} & Online & 54.9 & 24.4  \\
        & SG{-}Nav~\cite{sg-nav}              & Online & 54.0 & 24.9 \\
        & TriHelper~\cite{zhang2024trihelper} & Online &  56.5 & 25.3    \\
        & \TBD~(Ours)                         & Offline & \textbf{61.7} & \textbf{34.3} \\
            \midrule
    	\bottomrule
    \end{tabularx}
\end{table}

\subsection{Overall Performance Evaluation}

Table~\ref{table1:overall} reports the overall performance on the HM3D validation set. \TBD~achieves state-of-the-art results in both SR and SPL, demonstrating strong accuracy and efficiency. In particular, it surpasses ApexNav, a top-performing method that does not rely on online LLM queries, on both metrics and shows especially large gains of 3.52\% increase in SR. ApexNav neither incorporates contextual room cues nor adaptively emphasizes different types of cues based on the characteristics of the target, causing the agent to overlook critical semantic cues and make inaccurate and inefficient movements during exploration. In contrast, our approach adaptively weights different types of contextual cues, prioritizing the most informative semantic signals for each target and enabling accurate, robust navigation. \TBD~also outperforms methods that employ online LLM queries, most notably in SPL, demonstrating robust reasoning while maintaining efficient navigation. By extracting the most relevant contextual information about the target using commonsense knowledge prior to execution, our method enables the agent to perform high-level reasoning and actions efficiently and consistently.



\setlength{\tabcolsep}{11pt}
\begin{table}[t!]
    \captionsetup{font=footnotesize}
    \centering
    \caption{
    Ablation study of \TBD~modules on the HM3D dataset. We analyze the effect of incorporating contextual rooms and contextual objects, along with adaptive weighting, measured in terms of SR and SPL. Bold indicates the best performance.
    }
    \label{tab:ablation_modules}
    \begin{tabularx}
        {\linewidth}{ccccc}
        \toprule
        \midrule 
        \makecell{Contextual \\ rooms} &
        \makecell{Contextual \\ objects} &
        \makecell{Adaptive \\ weighting} &
        \makecell{SR↑} & 
        \makecell{SPL↑} \\ 
        \cmidrule(l{0.33em}r{.33em}){1-3} 
        \cmidrule(l{0.33em}r{.33em}){4-5}
        & & & 59.6 & 33.0 \\
        \checkmark & & &59.8 & 33.2\\
        & \checkmark & & 60.2& 31.8\\
        \checkmark & \checkmark & &61.2 &33.8 \\
        \checkmark & \checkmark & \checkmark & \textbf{61.7 }& \textbf{34.3} \\
        \midrule
        \bottomrule
    \end{tabularx}
\end{table}



\noindent These results confirm that our approach effectively unifies contextual global-level and local-level cues with offline commonsense knowledge, enabling a state-of-the-art ZSON that is both accurate and practical for real-world deployment.



\begin{figure*}[!t]
    \centering
    \captionsetup{font=footnotesize}    \includegraphics[width=1.0\linewidth]{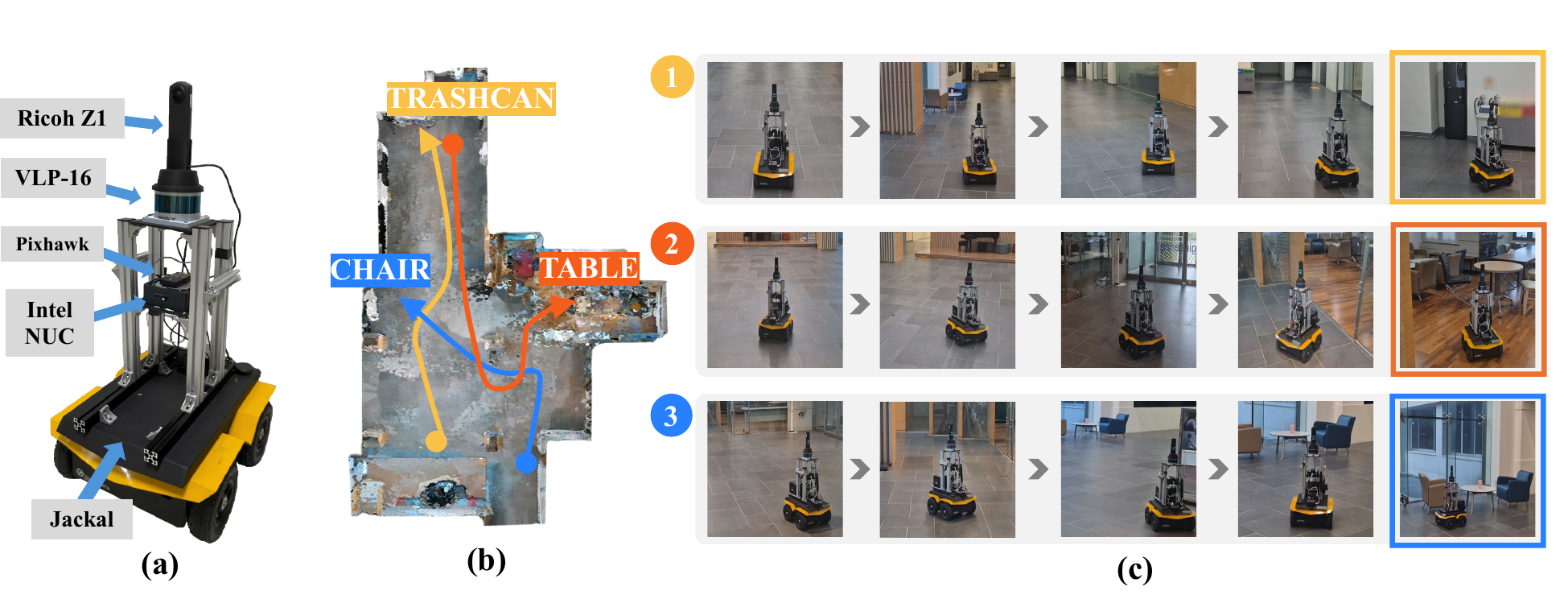}
    \caption{
    Configuration for real-world experiments. (a) A customized UGV platform based on a Clearpath Jackal, equipped with an Intel NUC, a Velodyne VLP-16 LiDAR, and a Ricoh Theta Z1 camera. (b) Real-world environments along with the trajectories the robot traveled to search for various objects. (c) The robot leverages contextual cues to move to likely object locations, then verifies the findings using multi-view verification.
    }
    \label{fig:realworld}
\end{figure*}


\setlength{\tabcolsep}{4pt}
\begin{table}[t!]
    \captionsetup{font=footnotesize}
    \centering
    \caption{
    Ablation study of \TBD~adaptive weighting strategy on the HM3D dataset. We analyze the effect of varying weights $\omega_\text{room}$ and $\omega_\text{object}$ on objects with different entropy levels $H({\target})$, measured in terms of SR. Bold indicates the best performance.
    }
    \label{tab:ablation_entropy}
    \begin{tabularx}
        {\linewidth}{lccccc} 
        \toprule
        \midrule 
        \multirow{2}{*}{{\begin{tabular}{@{}c@{}} ~ \\[-5pt] $\target$ (Episode)  \end{tabular}}} & 
        \multirow{2}{*}{\begin{tabular}{@{}c@{}} ~ \\[-5pt] $H(\target)$ \end{tabular}} & 
        \multicolumn{4}{c}{SR by ($\omega_\text{room}$, $\omega_\text{object}$)}  \\
        \cmidrule(l{0.33em}r{0.33em}){3-6} 
         & & \TBD & (1.0,0.0) & (0.5,0.5) & (0.0,1.0) \\ 
        \midrule
        Toilet~(398)        & 0.043 & \textbf{72.9} & 72.6          & 63.8          & 72.4          \\
        Bed~(433)           & 0.124 & 57.7 & 57.2          & \textbf{59.1}          & 57.7         \\
        Couch~(376)         & 0.203 & \textbf{61.7} & 60.6          & \textbf{61.7}         & 61.4          \\
        TV~(281)            & 0.462 & 39.8 & 37.4          & \textbf{40.2}          & 38.4          \\
        Chair~(428)         & 0.883 & \textbf{73.3} & 69.0          & 73.1          & 72.7          \\
        Potted Plant~(84)   & 0.915 & \textbf{42.9} & 38.1          & 41.7          & \textbf{42.9}         \\
        \midrule
        Total (2000) & - & \textbf{61.7} & 59.8            & 60.2            & 61.2          \\
        \midrule
        \bottomrule
    \end{tabularx}
\vspace{-0.3cm}
\end{table}

\subsection{Ablation Study}

We demonstrate the effectiveness of our proposed method through an ablation study on the HM3D dataset, as shown in Table~\ref{tab:ablation_modules}. Without contextual rooms and contextual objects, the agent struggles to gather sufficient semantic information and visual evidence, resulting in the lowest performance on both metrics. Each contextual information enables the agent to accumulate richer semantic information for more reliable guidance. Adaptive weighting further enhances the reliability of the resulting unified semantic map by prioritizing the most relevant semantic cues based on the target object. These results confirm that our proposed method improves the agent’s ability to locate the target object.

We further analyze the effectiveness of our adaptive weighting strategy. Specifically, we replace adaptive weighting based on target–room association entropy $H({\target})$ with manually defined fixed weights for the contextual room and contextual object scores. Assigning a higher weight to the room score forces the agent to prioritize global semantic cues regardless of the target, whereas a higher weight on the object score biases it toward local semantic cues. As shown in Table~\ref{tab:ablation_entropy}, our adaptive weighting framework achieves the highest overall success rate across different types of target objects, demonstrating its effectiveness. For low-entropy objects with strong spatial associations, such as a toilet ($H(\target)=0.043$), the room-only model performs well, while the object-only model suffers performance degradation. In contrast, for high-entropy objects that can appear in multiple locations, such as a chair ($H(\target)=0.883$) and potted plant ($H(\target)=0.915$), relying solely on room cues leads to inefficient exploration, whereas object cues provide stronger guidance. The model that combines both cues with equal weights still underperforms compared with \TBD, confirming that non-adaptive integration is suboptimal.

These results demonstrate that \TBD’s adaptive strategy of dynamically adjusting the importance of global and local cues according to the object’s characteristics is more effective than fixed weighting. This validates the core idea of our work, showing that optimizing the exploration strategy according to object entropy plays a decisive role.



\begin{figure}[!t]
    \centering
    \captionsetup{font=footnotesize}
     \includegraphics[width=1.0\linewidth]{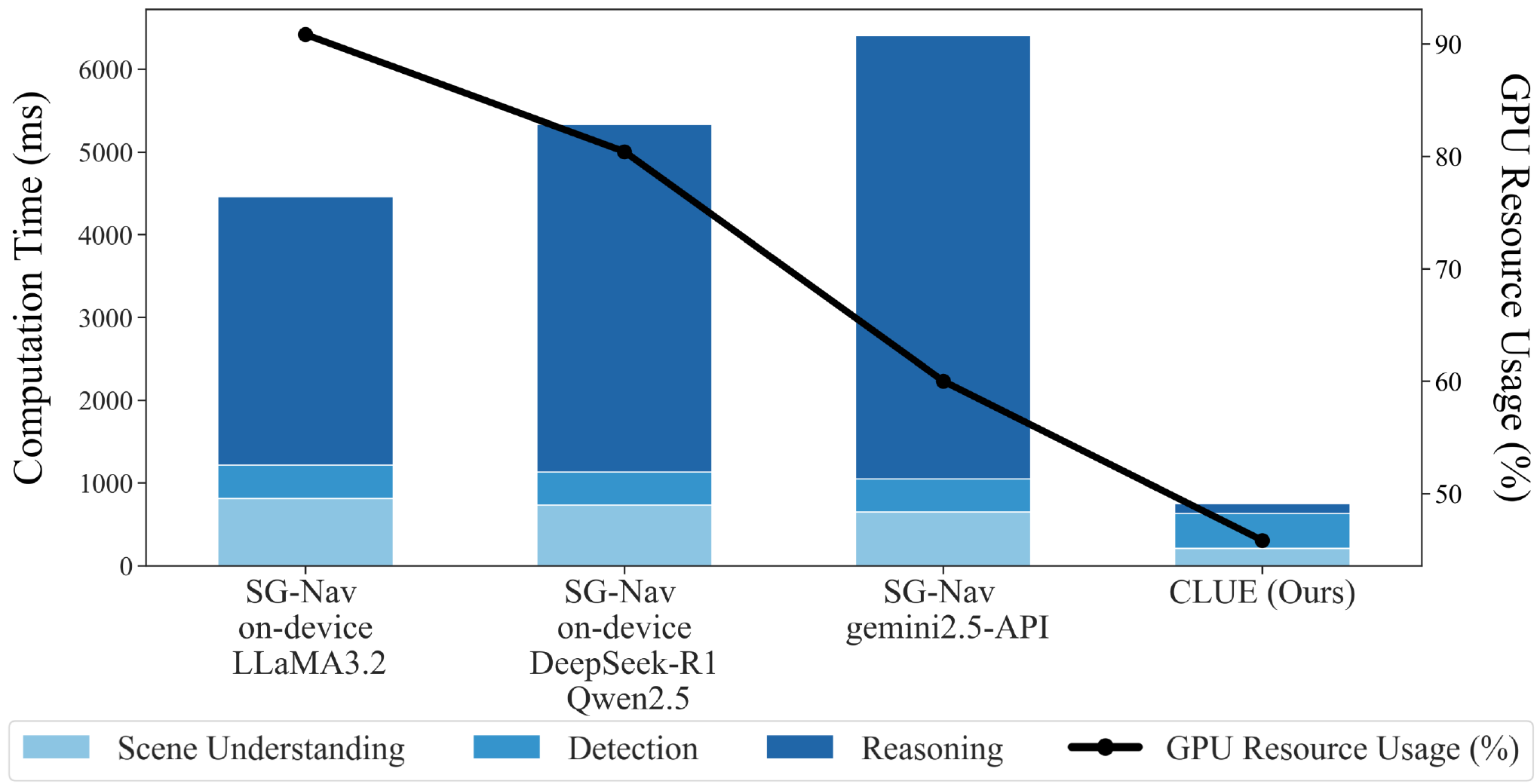}
    \caption{
    Resource usage analysis: comparison of computation time (ms) and GPU resource usage (\%) across different ObjectNav frameworks. These metrics are critical for evaluating performance on resource-constrained on-board systems.
    }
    \label{fig:resource}
\vspace{-15pt}
\end{figure}



\subsection{Resource Analysis}

We analyze the time consumption and GPU resource usage of \TBD~in comparison with competing baselines. In particular, we evaluate SG-Nav, which relies on online LLM queries with chain-of-thought reasoning and incremental prompting, techniques proposed to mitigate the high time complexity of LLM queries. We consider three variants of SG-Nav, distinguished by their model configurations: one with a single on-device LLaMA3.2-vision, a second with separate on-device  DeepSeek-R1(LLM) and Qwen2.5-VL(VLM) modules, and a third that leverages a gemini2.5-API LLM. To assess time consumption, we measure the per-step runtime for scene understanding, detection, and reasoning. For GPU usage, we report the peak memory required to execute each algorithm.

As shown in Fig.~\ref{fig:resource}, the on-device LLaMA3.2 model variant of SG-Nav consumes the most GPU resources, whereas on-device variants requires the most time for both scene understanding and decision-making. In contrast, \TBD~is considerably more efficient than SG-Nav, demanding fewer resources and delivering faster performance. It is also important to note that simulation benchmarks typically measure performance in terms of steps, often overlooking the actual runtime needed to complete an episode. This experiment demonstrates that \TBD~is well suited for navigation tasks on resource-constrained on-board systems.


\subsection{Real-World Experiment} 


To validate \TBD~in real-world settings, we conducted physical robot experiments. The robot configuration is shown in Fig.~\ref{fig:realworld}(a), and the test environment containing various objects is illustrated in Fig.~\ref{fig:realworld}(b). The system was evaluated using only text-based queries for different target objects. As shown in Fig.~\ref{fig:realworld}(c), our method successfully and consistently locates the specified objects, confirming its effectiveness in real-world scenarios. Notably, the system successfully identified targets even when they were placed in atypical locations, enabled by our unified semantic value map's comprehensive modeling of contextual cues and diverse correlations. In addition, multi-view verification further enhanced the robustness and reliability of object localization.



\section{Conclusion}
\label{sec:conclusion}

We presented CLUE, an adaptive framework for ZSON that prioritizes different types of contextual cues to enable effective and robust navigation. Unlike prior methods that apply uniform strategies or rely on costly online LLM queries, our approach adaptively prioritizes contextual cues and leverages offline commonsense knowledge to construct a unified semantic value map. This map integrates both contextual room and object cues, with their relative importance adaptively weighted by the target’s characteristics. Through this design, the agent balances global and local cues to infer likely object locations. Extensive experiments on the HM3D benchmark, along with real-world deployment on a Clearpath Jackal, show that \TBD~achieves superior accuracy and efficiency compared with state-of-the-art baselines. While effective, our framework assumes predefined room categories and relies on precomputed commonsense knowledge, which may limit generalization. Future work will extend the approach to an open-set setting and incorporate online adaptation for greater robustness in diverse scenarios.







\bibliographystyle{URL-IEEEtrans}
\bibliography{URL-bib}

\end{document}